\documentclass[10pt,twocolumn,letterpaper]{article}

\usepackage[pagenumbers]{iccv} 

\definecolor{iccvblue}{rgb}{0.21,0.49,0.74}
\usepackage[pagebackref,breaklinks,colorlinks,allcolors=iccvblue]{hyperref}

\def\paperID{} 
\def\confName{ICCV}
\def\confYear{2025}

\title{Shushing! Let's Imagine an Authentic Speech from the Silent Video}

\author{Jiaxin Ye, Hongming Shan\thanks{Corresponding author}\\
Fudan University, Shanghai, China\qquad\\
{\tt\small jxye22@m.fudan.edu.cn,\, hmshan@fudan.edu.cn}
}

\usepackage{amsmath}
\usepackage{amssymb}
\usepackage{bm}
\usepackage{url}
\usepackage{ragged2e}
\usepackage{multirow}
\usepackage{tablefootnote}
\usepackage{xspace}
\usepackage{xcolor,colortbl}
\usepackage{wrapfig}
\usepackage{makecell}
\usepackage[thicklines]{cancel}
\usepackage{xcolor}

\usepackage{tcolorbox}
\usepackage{bbding}
\usepackage{pifont}
\usepackage{arydshln} 
\usepackage{lipsum}
\usepackage{float}
\usepackage{booktabs}
\usepackage{utfsym}
\usepackage{subcaption}

\makeatletter

\def\@IEEEsectpunct{.\ \,}
\def\paragraph{\@startsection{paragraph}{4}{\z@}{1.0ex plus 1.ex minus 0.5ex}%
{0ex}{\bfseries}}%
\makeatother

\newcommand{\methodname}{ImaginTalk\xspace}
\newcommand{\taskname}{CV2S\xspace}

\usepackage{enumitem}

\definecolor{maroon}{rgb}{0,0,0}

\usepackage{bm}
\DeclareMathAlphabet{\mathsfit}{\encodingdefault}{\sfdefault}{m}{sl}
\SetMathAlphabet{\mathsfit}{bold}{\encodingdefault}{\sfdefault}{bx}{n}

\def\mP{{\bm{P}}}
\def\mQ{{\bm{Q}}}

\makeatletter

\DeclareRobustCommand{\cev}[1]{%
  {\mathpalette\do@cev{#1}}%
}
\newcommand{\do@cev}[2]{%
  \vbox{\offinterlineskip
    \sbox\z@{$\m@th#1 x$}%
    \ialign{##\cr
      \hidewidth\reflectbox{$\m@th#1\vec{}\mkern4mu$}\hidewidth\cr
      \noalign{\kern-\ht\z@}
      $\m@th#1#2$\cr
    }%
  }%
}

\begin{document}
\maketitle

\begin{abstract}

Vision-guided speech generation aims to produce authentic speech from facial appearance or lip motions without relying on auditory signals, offering significant potential for applications such as dubbing in filmmaking and assisting individuals with aphonia. 
Despite recent progress, existing methods struggle to achieve unified cross-modal alignment across semantics, timbre, and emotional prosody from visual cues, prompting us to propose \emph{Consistent Video-to-Speech (CV2S)} as an extended task to enhance cross-modal consistency. 
To tackle emerging challenges, we introduce \textbf{\methodname}, a novel cross-modal diffusion framework that generates faithful speech using only visual input, operating within a discrete space. 
Specifically, we propose a discrete lip aligner that predicts discrete speech tokens from lip videos to capture 
semantic information, while an error detector identifies misaligned tokens, which are subsequently refined through masked language modeling with BERT.
To further enhance the expressiveness of the generated speech, we develop a style diffusion transformer equipped with a face-style adapter that adaptively customizes identity and prosody dynamics across both the channel and temporal dimensions while ensuring synchronization with lip-aware semantic features.
Extensive experiments demonstrate that \methodname can generate high-fidelity speech with more accurate semantic details and greater expressiveness in timbre and emotion compared to state-of-the-art baselines. Demos are shown at our project page: \href{https://imagintalk.github.io}{https://imagintalk.github.io}.

\end{abstract}

\section{Introduction}
\label{sec:intro}

\begin{figure}[t]
    \centering
    \includegraphics[width=1.0\linewidth]{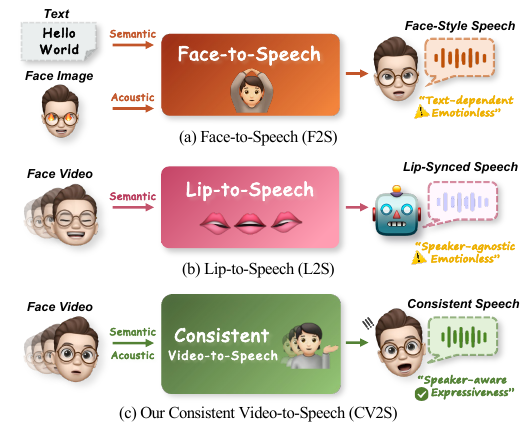}
    \caption{
    \textbf{Comparison of speech generation tasks.} (a) Face-to-Speech (F2S) generates speech from text and a face image, learning semantic and acoustic features, but remains dependent on text input and lacks emotional expressiveness. (b) Lip-to-Speech (L2S) generates speech solely from a face video, capturing semantic content but struggling to model the full range of acoustic features. (c) Our proposed Consistent Video-to-Speech (CV2S) generates speech solely from a face video, aligning with semantic content, facial identity, and emotional expression, presenting a novel approach to vision-guided speech generation.
    }
    \label{fig:task}
    \vspace{-4mm}
\end{figure}

Vision-guided speech generation aims to emulate the human ability to imagine and synthesize speech, generating authentic vocal expressions solely from facial appearance or lip motions without relying on auditory signals. 
This capability has significant real-world applications, including dubbing in filmmaking, assisting individuals with aphonia, and enabling fluent dialogue in silent-required environments. 

As interest in  vision-guided speech generation continues to grow~\cite{diffv2s:conf/iccv/ChoiHR23,Face2Speech:conf/interspeech/GotoOSTM20,facespeak:conf/cvpr/JangKAKYJKKC24,visualvoicecloning/ChenTQZLW22}, existing approaches can be broadly categorized into two paradigyms: Face-to-Speech (F2S) and Lip-to-Speech (L2S), depending on how  visual cues are processed for speech generation. 
As illustrated in Fig.~\ref{fig:task}(a) and (b),  F2S  aims to generate speech from an input text and facial image, ensuring that identity features such as timbre and tone are consistent with the provided face~\cite{facestylespeech:journals/corr/abs-2311-05844,facespeak:conf/cvpr/JangKAKYJKKC24,FaceTTS:conf/icassp/LeeCC23,demoface/abs-2502-01046}, while L2S focuses on learning a direct mapping from lip motions to the corresponding semantic information of speech~\cite{lip2speech_AAAI:conf/aaai/KimKC24,VCAGAN_lip:conf/nips/KimHR21,lipvoicer:conf/iclr/YeminiSBGF24,MTL_lip:conf/icassp/KimHR23}. 
However, both paradigms face challenges in achieving unified cross-modal speech synthesis. 
Specifically,  F2S learns face-style acoustic information from vision input and semantics from text input, while neglecting the semantic information inherent in visual cues (\ie,~lip motions). Conversely,  L2S focuses on phoneme generation from lip motions but struggles to model diverse vocal styles and emotional prosody. 

To bridge this gap, we introduce \emph{Consistent Video-to-Speech (\taskname)} as an extension of both tasks, aiming to achieve unified cross-modal alignment across semantics, timbre, and emotional prosody solely from visual cues. As illustrated in Fig.~\ref{fig:task}(c),  
\taskname converts silent video into speech by disentangling content, identity, and emotion from facial inputs, allowing fine-grained speech customization of both semantic and acoustic features.  
While CV2S may appear as a combination of F2S and L2S, it introduces additional challenges for existing methods as follows. (\textbf{i}) \emph{Insufficient cross-modal alignment}: Conventional F2S and V2S methods~\cite{facestylespeech:journals/corr/abs-2311-05844,VCAGAN_lip:conf/nips/KimHR21,SVTS:conf/interspeech/MiraHPSP22} primarily focus on either identity-alignment or lip-synchronization but fail to capture both implicit semantic and rich acoustic cues embedded in the visual inputs; 
and (\textbf{ii}) \emph{Dynamic prosody modeling}: Due to the temporal variability in videos, existing methods~\cite{FaceTTS:conf/icassp/LeeCC23,diffv2s:conf/iccv/ChoiHR23,demoface/abs-2502-01046} fail to model prosodic fluctuations, as they rely on a static identity embedding rather than learning dynamic speech modulation over time.

To tackle these challenges, we propose \emph{\methodname}, a novel cross-modal diffusion framework designed to generate authentic speech from silent talking videos.  \methodname is the first method to jointly model semantic, timbre, and emotional prosody solely from visual cues. 
Specifically, to mitigate the one-to-many mapping issues caused by the strong temporal and spectral correlations of continuous speech features~\cite{foundationTTS:journals/corr/abs-2303-02939}, we first formulate speech generation as a discrete token diffusion process using Residual Vector Quantization (RVQ).
Second, we design a discrete lip aligner that tokenizes cropped lip videos into vq-wav2vec tokens~\cite{vqwav2vec:conf/iclr/BaevskiSA20} for semantic information extraction. The lip aligner also consists of an error detector to identify misaligned tokens that are replaced with \texttt{[MASK]} token and refined by RoBERTa~\cite{roberta:journals/corr/abs-1907-11692} through strong masked language modeling with contextual information. 
Third, to enhance acoustic perception, we develop a style diffusion transformer with a face-style adapter, allowing adaptive customization of timbre and prosody dynamics across both channel and temporal dimensions. 
Extensive experiments demonstrate that \methodname achieves high-fidelity and expressive speech generation, outperforming state-of-the-art text-free vision-guided speech generation methods in consistency and intelligibility. 

Our contributions are summarized as follows.
\begin{itemize}[leftmargin=12pt]
\item We introduce Consistent Video-to-Speech (\taskname) as an extension to the existing vision-guided speech generation tasks,  addressing a critical gap in generating high-fidelity and diverse speech solely from vision cues.
\item We propose a novel cross-modal diffusion framework in the discrete space, mitigating one-to-many mapping issues to generate highly diverse and faithful speech.
\item We develop a style diffusion transformer with dual-level adaptive instance layer normalization, allowing modeling global timbre and temporal prosody across two dimensions.
\item Extensive experiments demonstrate that \methodname outperforms existing methods in semantic consistency, speech diversity, and prosodic expressiveness.
\end{itemize}

\section{Related Work}
\label{sec:relatedwork}

\paragraph{Face-to-speech (F2S) generation.}\quad
F2S aims to synthesize speech with reasonable timbre based on the speaker's visual appearance~\cite{FaceTTS:conf/icassp/LeeCC23,facespeak:conf/cvpr/JangKAKYJKKC24,facestylespeech:journals/corr/abs-2311-05844}.
Existing F2S methods have explored a promising solution by learning visual identity representation under speech guidance. For example, Lee~\etal~\cite{FaceTTS:conf/icassp/LeeCC23} and Goto~\etal~\cite{Face2Speech:conf/interspeech/GotoOSTM20} both introduce cross-modal alignment to minimize the distance between visual embedding and vocal speaker embedding, while these methods ignore the rich emotional cues inherent in the face. They often generate speech without diverse emotional prosody.
Although Ye~\etal~\cite{demoface/abs-2502-01046} explicitly learn utterance-level emotion embedding to enhance prosody modeling, they neglect dynamic variations and fail to achieve cross-modal alignment on underlying visual semantic information. 
Given that, these methods still struggle to address the challenges posed by \taskname task. In contrast, our \methodname is text-free, which can synthesize speech with dynamic emotional prosody and semantic information aligned with lip motions.

\paragraph{Lip-to-speech (L2S) generation.}\quad
L2S methods focus on reconstructing underlying speech aligning with lip motions from a silent talking video~\cite{lipvoicer:conf/iclr/YeminiSBGF24,diffv2s:conf/iccv/ChoiHR23,lip2speech_AAAI:conf/aaai/KimKC24,MTL_lip:conf/icassp/KimHR23,VCAGAN_lip:conf/nips/KimHR21}. Existing L2S methods primarily introduce a video encoder to predict target mel-spectrogram directly. 
For example, Yemini~\etal~\cite{lipvoicer:conf/iclr/YeminiSBGF24} develop a diffusion model conditioned on visual representations to predict mel-spectrogram and conduct classifier guidance using a lip reading expert.
However, we observe that the generated speech of existing methods still suffers from low quality and, including over-smoothed acoustic details, inaccurate timbre, and unnatural prosody. The limitation may arise from the one-to-many mapping issues caused by continuous speech features, as they are highly correlated along time and frequency dimensions, leading to over-smoothing during frames for these models to predict~\cite{foundationTTS:journals/corr/abs-2303-02939}. 
In this work, our \methodname proposes to learn speech features in the discrete space that preserve rich semantic and acoustic information, allowing speech generation with high fidelity and diversity. 

\paragraph{Emotional speech generation.}\quad Emotional speech generation seeks to enrich generated speech with emotional prosody~\cite{mmtts_emo:journals/corr/abs-2404-18398,visualvoicecloning/ChenTQZLW22,emodiff:conf/icassp/GuoDCY23}. 
Existing methods can be broadly divided into two categories based on how to integrate emotion information into the generation process. 
For emotion label conditioning, EmoDiff~\cite{emodiff:conf/icassp/GuoDCY23} introduces a diffusion model with soft emotion labels using classifier guidance. 
For emotional representation learning, V2C-Net~\cite{visualvoicecloning/ChenTQZLW22} and HPM~\cite{visualvoicecloning/Cong0QZWWJ0H23} learn emotion and speaker representations from reference face and speech individually for speech customization. 
However, previous methods primarily focus on learning utterance-level (\ie, global-level) emotion features to guide the speech generation, neglecting dynamic variations in prosody modeling. Our \methodname proposes a dual-level adaptive instance normalization to synthesize speech with dynamic emotional prosody.

\section{Preliminary: Discrete Diffusion Models}
\label{subsec:discrete_diffusion}

Recently, continuous diffusion models (CDM)~\cite{fastdiff:conf/ijcai/HuangL0S00Z22,mixganTTS:journals/access/DengWQLC23,diffv2s:conf/iccv/ChoiHR23,emodiff:conf/icassp/GuoDCY23} have achieved state-of-the-art results in speech generation, while they are limited by one-to-many issues and computational inefficiency, frustrating practical application. An intuitive solution is to utilize discrete speech tokens~\cite{encodec:journals/tmlr/DefossezCSA23,Srcodec:conf/icassp/ZhengTXX24,maskgct:journals/corr/abs-2409-00750} to build discrete diffusion models (DDMs), which have shown promising in language modeling~\cite{D3PM:conf/nips/AustinJHTB21,ConcreteScoreMatch:conf/nips/MengCSE22,SEDD:conf/icml/LouME24} and speech generation~\cite{diffsound:journals/taslp/YangYWWWZY23,DCTTS:conf/icassp/WuLLY24,demoface/abs-2502-01046}. 
In this paper, we introduce the DDM to generate speech tokens based on cross-modal guidance, and outline below the forward and reverse processes of the DDM, along with its training objective.

\paragraph{Forward diffusion process.}\quad
Given a token sequence $\bm{x} = [x^1, \ldots, x^d]$ with length $d$, where each token belongs to a discrete state sapce $\mathcal{X} = \{1, \ldots, n\}$. The diffusion process can be modeled as a continuous-time discrete Markov chain, parameterized by the diffusion matrix $\mQ_t\in \mathbb{R}^{n^d\times n^d}$, also known as the transition rate matrix at time $t$, as follows:
\begin{equation}
\label{eq:delta_transition}
    p({x}_{t+ \Delta t}^i|x_t^i) =  \delta_{{x}^i_{t+ \Delta t}x^i_t} + \mQ_t({x}^i_{t+ \Delta t},x^i_t)\Delta t + o(\Delta t),
\end{equation}
where $\delta$ is Kronecker delta, $x^i_t$ denotes $i$-th element of $\bm{x}_t$, and $\mQ_t({x}^i_{t+ \Delta t},x^i_t)$ is the $({x}^i_{t+ \Delta t},x^i_t)$ element of $\mQ_t$, which represents the transition rate from state $x^i_t$ to state ${x}^i_{t+ \Delta t}$ at time $t$. To further achieve efficient computation, existing methods~\cite{SEDD:conf/icml/LouME24,RADD:journals/corr/abs-2406-03736} adopt the assumption of dimensional independence, conducting a one-dimensional diffusion process for each dimension with the same token-level diffusion matrix $\mQ_t^\text{tok}=\sigma(t)\mQ^\text{tok}\in \mathbb{R}^{n\times n}$, where $\sigma(t)$ is the noise schedule and $\mQ^\text{tok}$ is designed to diffuse towards an masked state \texttt{[MASK]}. 
Now, the forward equation can be formulated as $\mP({x}^i_t,x^i_0) = \exp\left(\bar{\sigma}(t) \mQ^\text{tok}({x}^i_t,x^i_0) \right)$, where transition probability matrix $\mP({x}^i_t,x^i_0) := p({x}^i_t|x_0)$, and cumulative noise $\bar{\sigma}(t) = \int_0^t \sigma(s)ds$. There are two probabilities in the $\mP_{t|0}$: \( 1 - e^{-\bar{\sigma}(t)} \) for replacing the current tokens with \texttt{[MASK]}, \( e^{-\bar{\sigma}(t)} \) for remaining unchanged. 
Finally, the corrupted sequence $\bm{x}_t$ can be sampled from $\bm{x}_0$ in one step.

\paragraph{Reverse denoising process.}\quad
Given the diffusion matrix $\mQ^\text{tok}_t$, we need a reverse transition rate matrix $\bar{\mQ}_t$~\cite{SCDDM:conf/iclr/SunYDSD23,kelly2011reversibility} to formulate reverse process, where $\bar{\mQ}_t(x^i_{t- \Delta t},x^i_t)=\tfrac{ p(x^i_{t- \Delta t})}{p(x^i_t)} \mQ^\text{tok}_t(x^i_t,x^i_{t- \Delta t})$ and $x^i_{t- \Delta t}\neq x^i_t$, or $\bar{\mQ}_t(x^i_{t- \Delta t}, x^i_t) =  - \sum_{z \neq x_t} \bar{\mQ}_t(z,x^i_t)$. 
The reverse equation is formulated as follows: 
\begin{equation}
    \label{eq:backward}
    p({x}^i_{t- \Delta t}|x^i_t) =  \delta_{{x}^i_{t- \Delta t}x^i_t} + \bar{\mQ}_t({x}^i_{t- \Delta t},x^i_t)\Delta t + o(\Delta t).\\
\end{equation}
The core of the reverse denoising process is to estimate the concrete score $c_{x^i_{t- \Delta t} {x}^i_t}=\tfrac{p(x^i_{t- \Delta t})}{p(x^i_t)}$ of $\bar{\mQ}_t$, representing to measure the \textit{transition probability or closeness} from a state $x^i$ at time $t$  to a state $\hat{x}^i$ at time $t- \Delta t$. We can introduce a score network $s_\theta({x}^i_t,t)_{x^i_{t- \Delta t}} \approx [\tfrac{p(x^i_{t- \Delta t})}{p(x^i_t)}]_{x^i_{t}\neq x^i_{t- \Delta t}}$ to learn the score, so that the reverse matrix is parameterized to model the reverse process $q_\theta({x}^i_{t- \Delta t}|x^i_t)$ (\ie,~parameterize the concrete score).

\paragraph{Training objective.~~\xspace}
Denoising score entropy (DSE)~\cite{SEDD:conf/icml/LouME24} is introduced to train the score network $s_\theta$:
\begin{equation}
\label{eq:score_entropy}
\begin{aligned}
       \int_0^T \mathbb{E}_{\bm{x}_t \sim p\left(\bm{x}_t \mid \bm{x}_0\right)} \sum_{{\hat{\bm{x}}_t} \neq \bm{x}_t} \mQ_t\left( \hat{{x}}^i_t,{x}^i_t\right)  \Big[s_\theta\left({x}^i_t, t\right)_{\hat{{x}}^i_t}  & \\
        - c_{\hat{{x}}^i_t {x}^i_t} \log s_\theta\left({x}^i_t, t\right)_{\hat{{x}}^i_t}+  \text{N}(c_{\hat{{x}}^i_t {x}^i_t})\Big] dt &,
\end{aligned}
\end{equation}
where the concrete score $c_{\hat{{x}}^i_t {x}^i_t} = \tfrac{p\left({\hat{{x}}^i_t} \mid {x}^i_0\right)}{p\left({x}^i_t \mid {x}^i_0\right)}$ and a normalizing constant function $\text{N}(c):= c \log c - c$ that ensures loss non-negative. 
During sampling, we start from $\bm{x}_{T}$ filled with masked token \texttt{[MASK]}, and iteratively sample new set of tokens $\bm{x}_{t-1}$ from $p_{\theta}(\bm{x}_{t-1}|\bm{x}_{t})$ by replacing the concrete score with the trained score network on~\cref{eq:backward}.

\section{Methodology}
\label{sec:method}
In this section, we describe our \methodname, a novel cross-modal diffusion framework to achieve consistent video-to-speech generation. 
We present the task formulation in Sec.~\ref{subsec:task} about \taskname, an overview of \methodname in Sec.~\ref{subsec:overview}, and then detail its key components. 

\begin{figure*}[t]
    \centering
    \includegraphics[width=1.0\linewidth]{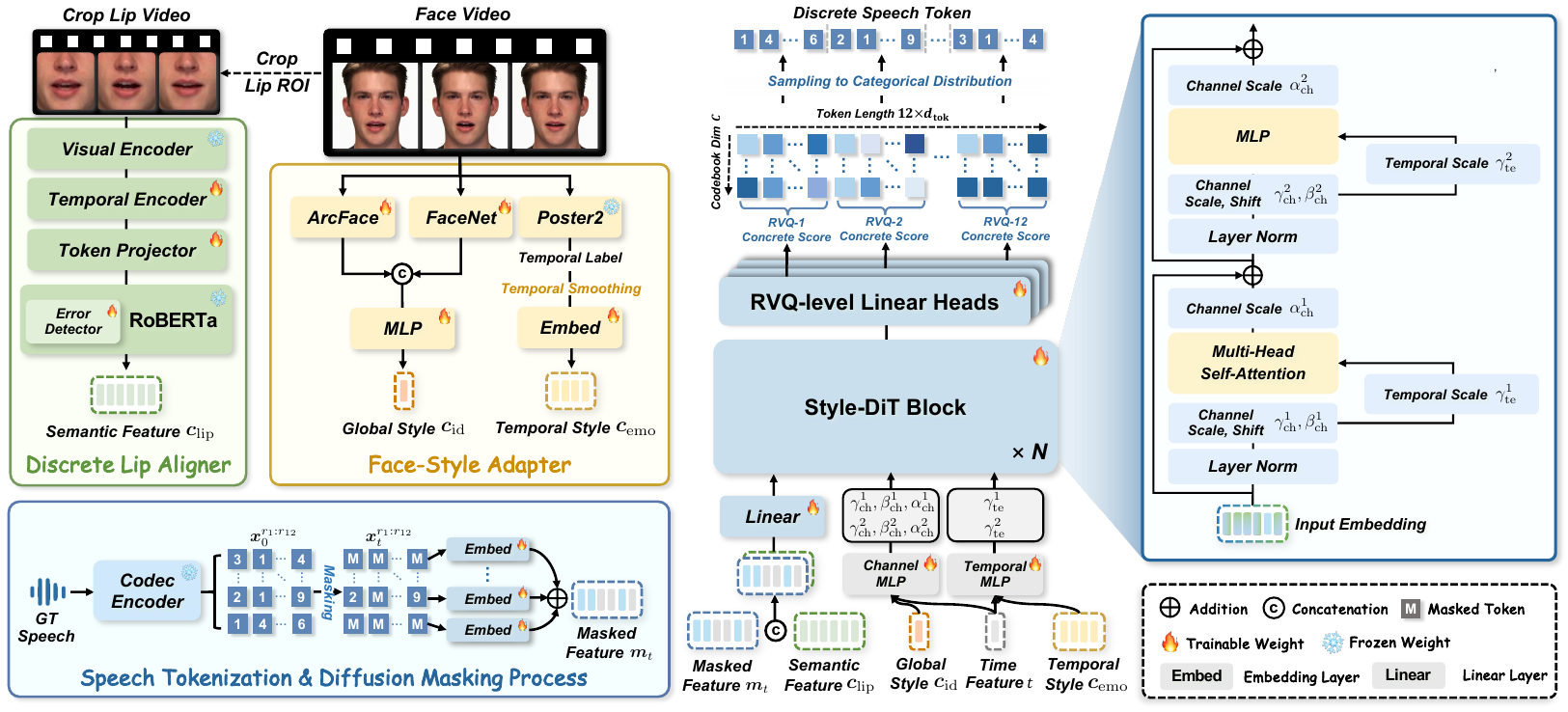}
    \caption{\textbf{Overall framework of \methodname}. The input face video is first processed by the face-style adapter to extract the global identity style $\bm{c}_\text{id}$ and temporal prosody style $\bm{c}_\text{emo}$, while the lip region of interest (ROI) is cropped and processed by a discrete lip aligner to learn semantic features $\bm{c}_\text{lip}$. Furthermore, masked speech features are obtained via the codec encoder and forward diffusion process. 
    Finally, the Style-DiT takes these features as inputs to predict concrete scores through \( N \) Style-DiT blocks and 12 linear heads for 12 level tokens. }
    \label{fig:main}
\end{figure*}

\subsection{Task Formulation for \taskname}
\label{subsec:task}
Given a silent talking video $\mathcal{V}$, the \taskname task aims to reconstruct speech that aligns with the lip motions, facial appearance, and emotional expression (\ie,~semantic content, timbre, and emotional prosody), all from only a video $\mathcal{V}$. 
More precisely, it would like to extract a triplet of visual conditions $\bm{c} = \{\bm{c}_\text{lip}, \bm{c}_\text{id}, \bm{c}_\text{emo}\}$, which correspond to reference lip motions, facial identity, and emotional expression. The synthesized speech content should align with the lip motions $\bm{c}_\text{lip}$, while its voice identity and emotional prosody attributes correspond to the identity $\bm{c}_\text{id}$ and emotion $\bm{c}_\text{emo}$, all extracted from the input video.

\subsection{Overview of \methodname}
\label{subsec:overview}
Fig.~\ref{fig:main} illustrates the overview of \methodname. 
First, to ensure cross-modal alignment, we propose a discrete lip aligner and a face-style adapter to capture semantic and acoustic style information. The face-style adapter extracts the global identity style $\bm{c}_\text{id} \in \mathbb{R}^{C_\text{id}}$ and temporal prosody style $\bm{c}_\text{emo}\in \mathbb{R}^{L' \times C_\text{emo}}$ from the input video, while the lip region of interest (ROI) is cropped and processed by a discrete lip aligner to learn semantic features $\bm{c}_\text{lip}\in \mathbb{R}^{L\times C}$, where $C, C_\text{id}$ and $C_\text{emo}$ are channel sizes, and $L, L'$ denote lengths
Additionally, we utilize the codec encoder to extract RVQ speech tokens and conduct the forward diffusion process as~\eqref{eq:delta_transition} to obtain masked speech feature $\bm{m}_t\in \mathbb{R}^{L\times C}$. 
Building upon these features, the Style-DiT inputs the masked feature $\bm{m}_t$ concatenated with $\bm{c}_\text{lip}$ in the channel dimension for the semantic alignment, while scale and shift parameters of our Dual-level adaptive Layer Norm (Dual-adaLN) serve as additional inputs to inject acoustic information. 

Next, we detail the key components in \methodname.

\subsection{Vision-guided Semantic and Acoustic Modeling}
\paragraph{Discrete lip aligner for semantic modeling.}\quad
Inspired by the video-speech recognition method SyncVSR~\cite{syncvsr:journals/corr/abs-2406-12233}, which maps lip video into discrete tokens, we develop a discrete lip aligner with RoBERTa~\cite{roberta:journals/corr/abs-1907-11692} to learn the semantic features $\bm{c}_\text{lip}\in \mathbb{R}^{L\times C}$ from lip motions. As shown in Fig.~\ref{fig:main}, the lip aligner consists of a visual encoder, a temporal encoder, and a token projector. The visual encoder utilizes a ResNet architecture~\cite{resnet} with 3D convolution to capture both local features and contextual information. The temporal encoder, based on a conformer~\cite{conformer}, models long-term semantic dependencies, followed by a linear layer that acts as the token projector to predict vq-wav2vec tokens~\cite{vqwav2vec:conf/iclr/BaevskiSA20}. 
Moreover, to mitigate the influence of misaligned tokens in cross-modal alignment, we introduce an error detector that marks tokens with a cosine similarity to the ground truth embedding below 0.9 as misaligned. During inference, these tokens are replaced with the $\texttt{[MASK]}$ token and re-extracted by RoBERTa via masked language modeling using contextual information. During training, we directly obtain the semantic feature $\bm{c}_\text{lip}$ from the target vq-wav2vec tokens.

\paragraph{Face-style adapter for acoustic modeling.}\quad
For acoustic modeling on timbre and prosody, the timbre can reflect an individual's identity as a global speech style, while prosody captures temporal dynamics, typically conveying emotion. 
As presented in Fig.~\ref{fig:main}, we build a encoder composing with two face recognition models ArcFace~\cite{arcface:journals/pami/DengGYXKZ22} and FaceNet~\cite{facenet:conf/cvpr/SchroffKP15} to learn global style $\bm{c}_\text{id}\in \mathbb{R}^{C_\text{id}}$ under supervision of a strong speaker recognition model GE2E~\cite{GE2E/WanWPL18}. Specifically, we desire to align $\bm{c}\text{id}$ with the embedding of the GE2E by minimizing cosine similarity, $\ell_1$, and $\ell_2$ losses. 
Additionally, we introduce a strong video facial expression recognition model Poster2~\cite{posterv2:journals/corr/abs-2301-12149}. To decouple identity information, we first predict the emotional class for total $L$ frames and apply a temporal smoothing strategy to average logits within a 0.5-second window, yielding a subsequence of length $L'$. Then, we utilize a learnable embedding layer to extract identity-agnostic temporal features $\bm{c}_\text{emo}\in \mathbb{R}^{L'\times C_\text{emo}}$. More details about training can be refereed in Sec.~\ref{subsec:training}.

\subsection{Diffusion-based Speech Token Generation}
\label{subsec:f2a}
\paragraph{Speech tokenization.}\quad
Given a single-channel speech signal, the the RVQ-based codec~\cite{maskgct:journals/corr/abs-2409-00750} compresses it into tokens represented as $\bm{x}^{r_1:r_{12}} = \{1,\ldots,C_\text{code}\}^{12\times d_\text{tok}}$, where $r_i$ is the $i$-th RVQ level of token, $d_\text{tok}$ is the length of the token sequence, respectively. The number of RVQ layers is 12 with a codebook size $C_\text{code}=1,024$ in each layer. 
 
\paragraph{Discrete diffusion process.}\quad
Following the diffusion process~\cite{SEDD:conf/icml/LouME24}, for the hierarchical structure of $\bm{x}^{r_1:r_{12}}$, we randomly mask each level token $\bm{x}^{r_i}_t$ at timestep $t$ into the masked state $[\texttt{MASK}]$. 
Specifically, we first extract input tokens  $\bm{x}^{r_1:r_{12}}_0$ from the codec encoder according to the curriculum training stage, where $r_l$ denotes the max level for the current input. After that, the diffusion process is conducted as defined in~\cref{eq:delta_transition} for $\bm{x}^{r_i}_t$, where $1\leq i\leq 12$.

\paragraph{Style-DiT with Dual-adaLN.}\quad 
The goal of our proposed Style-DiT is to respond faithfully to semantic and acoustic conditions, predicting concrete scores of each level speech tokens. 
Firstly, to achieve frame-level semantic alignment, the masked speech features $\bm{m}_t\in \mathbb{R}^{L\times C}$ is concatenated with $\bm{c}_\text{lip}$ along the channel dimension, and we utilize a linear layer for shape alignment. 
Moreover, different from class-driven DiT architecture~\cite{DiT:conf/iccv/PeeblesX23}, we propose a Dual-adaLN across channel and temporal dimensions to process different conditional inputs. The Dual-adaLN comprises two distinct parameter sets. The channel MLP predicts channel-level scale and shift parameters $\alpha_\text{ch}^1,\gamma_\text{ch}^1,\beta_\text{ch}^1,\alpha_\text{ch}^2,\gamma_\text{ch}^2,\beta_\text{ch}^2\in\mathbb{R}^{C}$ based on global style and time features, while the temporal MLP predicts temporal-level scale parameters $\gamma_\text{te}^1, \gamma_\text{te}^2\in\mathbb{R}^{L'}$ using temporal style and time features. As shown in Fig.~\ref{fig:main}, for the hidden input embedding $\bm{h}_t \in \mathbb{R}^{L\times C}$, we formulate the Dual-adaLN as:
\begin{equation}
\label{eq:duallevel}
\underbrace{\gamma_\text{te}^i\otimes \mathbf{1}_{25}}_\text{Temporal-level} \cdot \underbrace{\left((1+\gamma_\text{ch}^i)\cdot\frac{\bm{h}_t-\mu(\bm{h}_t)}{\sigma(\bm{h}_t)}+\beta_\text{ch}^i\right)}_\text{Channel-level},
\end{equation}
where $i=\{1,2\}$, $\otimes$ denotes Kronecker product, and $\mathbf{1}_{25}\in\mathbb{R}^{25}$ is an all-ones vector to up-sample $\gamma_\text{te}^i$ with $L'=\frac{L}{25}$ parameters to align with the hidden embedding with 50 Hz sampling rate. $\mu(\cdot)$ and $\sigma(\cdot)$ are the mean and standard deviation for $\bm{h}_t$ across the channel dimension. After employing the Dual-adaLN, we can inject the global style while capturing the temporal dynamics of emotional prosody, enabling a more faithful adaptation to various conditional inputs.

Finally, for the output, we incorporate 12 linear heads with a combination of Dual-AdaLN and a linear layer to predict concrete scores for each RVQ level.

\begin{table*}[thbp]
\setlength\tabcolsep{5.0pt}
\small
\centering
\begin{tabular}{rcccrrrrrrr}
\toprule
\multirow{2}{*}{Methods} & \multirow{2}{*}{Audio}  & \multirow{2}{*}{Text}  & \multirow{2}{*}{Vision}  & \multicolumn{3}{c}{\textit{Expressiveness}} & \multicolumn{2}{c}{\textit{Naturalness}}& \multicolumn{2}{c}{\textit{Synchronization}}\\
 &   &  &  & EmoSim$\uparrow$ & SpkSim$\uparrow$ & RMSE$\downarrow$ &MCD$\downarrow$  & WER(\%)$\downarrow$ & LSE-C$\uparrow$ & LSE-D$\downarrow$\\ \hline
\specialrule{0em}{3.5pt}{1.5pt}
Ground Truth & - & - & - & 1.00 & 1.00 & 0.00 & 0.00 & 5.90 & 3.33 & 8.15 \\ 
\hdashline
\specialrule{0em}{1.5pt}{1.5pt}
 \multicolumn{4}{c}{\textit{\textbf{Multimodal-to-Speech Generation}}} \\
V2C-Net~\cite{visualvoicecloning/ChenTQZLW22}  & \ding{51} & \textcolor[RGB]{127,127,127}{\ding{51}}  & \ding{51}  & 0.70 & 0.49  & 98.35 & 7.29  & \textcolor[RGB]{127,127,127}{25.54} & 2.11 & 9.57 \\ 
HPM~\cite{visualvoicecloning/Cong0QZWWJ0H23}  &\ding{51} &  \textcolor[RGB]{127,127,127}{\ding{51}}  & \ding{51} & 0.68 & 0.38  & 96.78 & 8.23  & \textcolor[RGB]{127,127,127}{59.58} & 2.51 & 9.07 \\  
\specialrule{0em}{1.pt}{1.5pt}
\hline
\specialrule{0em}{1.pt}{1.5pt}
\multicolumn{4}{c}{\textit{\textbf{Face-to-Speech Generation}}}\\
Face-TTS$^\dag$~\cite{FaceTTS:conf/icassp/LeeCC23}  &\ding{55} &  \textcolor[RGB]{127,127,127}{\ding{51}} & \ding{51}  & 0.52 & 0.19  & 113.62 & 9.52 & \textcolor[RGB]{127,127,127}{15.21} & 1.82 & 9.99 \\ 
\specialrule{0em}{1.pt}{0.5pt}
DEmoFace~\cite{demoface/abs-2502-01046} & \ding{55} & \textcolor[RGB]{127,127,127}{\ding{51}} & \ding{51}  & 0.73 & 0.58  & 96.62 & 7.48 & \textcolor[RGB]{127,127,127}{25.60} & 1.96 & 9.72 \\ 
\specialrule{0em}{1.pt}{1.5pt}
\hline
\specialrule{0em}{1.pt}{1.5pt}
 \multicolumn{4}{c}{\textit{\textbf{Lip-to-Speech Generation}}} \\
MTL$^\dag$~\cite{MTL_lip:conf/icassp/KimHR23} & \ding{51} & \ding{55} & \ding{51} & 0.52 & 0.35  & 115.84 & 9.63 & 109.59 & \textbf{3.66} & \textbf{7.75}\\ 
Lip2Wav$^\dag$~\cite{lip2wav:conf/cvpr/PrajwalMNJ20}  & \ding{51} & \ding{55}  & \ding{51}  & 0.57 & 0.44  & 89.31 & 12.58  & 132.55 & 2.77 & 8.71 \\ 
VCA-GAN$^\dag$~\cite{VCAGAN_lip:conf/nips/KimHR21}  & \ding{55} & \ding{55} & \ding{51} & 0.45 & $-$0.13  & 119.19 & 11.47 & 146.24 & 1.69 & 9.95 \\ 
LipVoicer$^\dag$~\cite{lipvoicer:conf/iclr/YeminiSBGF24}  &\ding{55} &  \ding{55}  & \ding{51} & 0.51 & 0.05  & 122.68 & 16.60  & 97.31& 2.72 & 8.70 \\ 
\specialrule{0em}{1.pt}{0.5pt}
\hline 
\specialrule{0em}{1.pt}{0.5pt}
 \methodname (\textbf{ours}) &  \ding{55} &  \ding{55} & \ding{51}  & \underline{0.74} & \underline{0.61}  & \underline{88.62} & \underline{7.73} & \textbf{64.66} & \underline{3.46} & \underline{8.03} \\ 
 ImaginTalk$^*$ (\textbf{ours})&  \ding{51} &  \ding{55} & \ding{51}  & \textbf{0.79} & \textbf{0.74}  & \textbf{86.32} & \textbf{7.37} & \underline{66.04} & 3.35 & 8.13 \\ 
\bottomrule
\end{tabular}
\caption{ \textbf{Quantitative results on unseen speakers.} The \textit{Audio}, \textit{Text}, and \textit{Vision} indicate whether specific modality conditions are used as guidance. $\uparrow$ ($\downarrow$) means the higher (lower) value is better. The best-performing method is \textbf{bolded}, while the second-best method is \underline{underlined}. Notably, $^*$ denotes that timbre conditioning is guided by speech features rather than the visually driven $\bm{c}_\text{id}$, $^\dag$ indicates that the results are obtained using the officially released model pre-trained on larger-scale datasets LRS3, and \textcolor[RGB]{127,127,127}{grey} indicates the additional text modality for semantic generation, excluded from WER ranking to avoid unfair semantic evaluation. \label{tab:main}}
\vspace{-2mm}
\end{table*}

\subsection{Training and Inference}
\label{subsec:training}
The training procedure for our \methodname contains three stages: semantic-lip alignment, acoustic-face alignment, and concrete score prediction. 
\paragraph{Training for semantic-lip alignment.}\quad
The lip-aligner is optimized with two loss terms to ensure synchronization between video and speech, including a cross-entropy loss $\mathcal{L}_\text{lip-token}$ to predict the quantized vq-wav2vec token $z_i$ from the input video $\mathcal{V}$ at time $i$, and a binary cross-entropy loss $\mathcal{L}_\text{lip-error}$ for identifying misaligned tokens. 
Specifically, the temporal encoder and token projector are optimized using $ \mathcal{L}_\text{lip-token} = -\frac{1}{L}\sum_{i\leq L}\log p(z_i\mid \mathcal{V})$. For misalignment detection, we first extract $\bm{e}_i^\text{pred}$, the predicted embedding of token $z_i$ from the RoBERTa, and the token with cosine similarity to the ground truth embedding $\bm{e}_i^\text{gt}$ below 0.9 is labeled as mis-aligned one (\ie,~label $y_i=0$). Then, the detector is optimized by $\mathcal{L}_\text{lip-error}= y_i  \log(p_i) + (1 - y_i)  \log(1 - p_i)$, where $y_i$ denotes aligned label and $p_i=p(\bm{e}_i^\text{pred})$ is the predicted probability of the detector. 

\paragraph{Training for acoustic-face alignment.}\quad 
For the global style $\bm{c}_\text{id}$, we introduce cosine similarity, $\ell_1$, and $\ell_2$ losses to align the global style vectors with speech GE2E vectors $\bm{c}_\text{GE2E}$, which the training objective for the ArcFace, FaceNet, and MLP is as $\mathcal{L}_\text{face-global} = 1-\mathrm{cos}(\bm{c}_\text{id},\bm{c}_\text{GE2E}) + \ell_1(\bm{c}_\text{id},\bm{c}_\text{GE2E})+\ell_2(\bm{c}_\text{id},\bm{c}_\text{GE2E})$. 
For the temporal style, we utilize the speech emotion recognition model~\cite{emo2vec:conf/acl/MaZYLGZ024} to obtain temporal labels of speech, shared with the same emotion class with the expression model Poster2. Then, the embedding layer is optimized with Style-DiT. 

\paragraph{Training for concrete score prediction.}\quad
The Style-DiT is optimized by multi-level DSE loss based on~\cref{eq:score_entropy} with the sum across all 12 RVQ levels as $\mathcal{L}_\text{score} = \sum_{i=1}^{12}\mathcal{L}_\text{DSE}(\bm{x}^{r_i},t,\bm{c})$. For conducting predictor-free guidance, we randomly set $\varnothing$ with 10\% probability for each condition and enforce all conditions set to $\varnothing$ for 10\% samples. 

\paragraph{Inference.}\quad
Following~\cref{eq:backward}, the reverse process is executed with Euler sampling~\cite{SEDD:conf/icml/LouME24} and enhanced predictor-free guidance~\cite{demoface/abs-2502-01046} with 64 steps. 
Notably, to avoid information degradation with teacher-student distillation, during training, we utilize ground truth semantic features, global style, and temporal style from the target speech instead of these features extracted from the video. During inference, based on pre-trained lip-aligner and face-adapter, we can extract relevant features solely from the video input for speech generation without other modalities.

\section{Experimental Results}
\label{sec:result}

\subsection{Experimental Setups}
\paragraph{Datasets.}\quad
All our models are pre-trained on two emotional datasets and a talk recordings dataset without emotion annotations: RAVDESS~\cite{RAVDESS} and MEAD~\cite{MEAD:conf/eccv/WangWSYWQHQL20} to enhance prosody modeling, and an additional 10-hour subset from LRS3~\cite{LRS3/abs-1809-00496} allowing our model to be comparable to lip-voicer~\cite{lipvoicer:conf/iclr/YeminiSBGF24} trained on 400 hours of LRS3. 
We randomly split the three datasets into train, validation, and test sets, ensuring no speaker overlap. 
Furthermore, we employ the SepFormer~\cite{sepformer/SubakanRCBZ21} model for speech enhancement, while Whisper~\cite{whisper/RadfordKXBMS23} is utilized to filter out misaligned text-speech pairs.
Finally, the pre-processing dataset comprises 29.02 hours of audio recordings and 24,089 utterances across 7 basic emotions (\ie~angry, disgust, fear, happy, neutral, sad, and surprised) and 789 speakers.

\paragraph{Evaluation metrics.}\quad
For \taskname, we evaluate the generation performance based on expressiveness, naturalness, and synchronization. 
For the expressiveness, we calculate cosine similarity metrics based on emotion embeddings~\cite{emo2vec:conf/acl/MaZYLGZ024} and d-vectors~\cite{dvector:journals/corr/LiWZZ15b} to obtain emotion similarity (EmoSim) and speaker identity similarity (SpkSim). Additionally, we evaluate the Root Mean Square Error (RMSE) for F0~\cite{RMSEf0:conf/asru/HayashiTKTT17}.
For the naturalness, we quantify spectral differences with Mel Cepstral Distortion (MCD)~\cite{visualvoicecloning/ChenTQZLW22}, as well as the Word Error Rate (WER)~\cite{WER/WangSZRBSXJRS18,whisper/RadfordKXBMS23} to gauge intelligibility. 
For the synchronization, we report the distance and confidence scores of lip sync errors (LSE-C and LSE-D) between speech and video using the pre-trained SyncNet~\cite{syncnet:conf/accv/ChungZ16a}.

\begin{figure*}[t]
    \centering
    \includegraphics[width=1.0\linewidth]{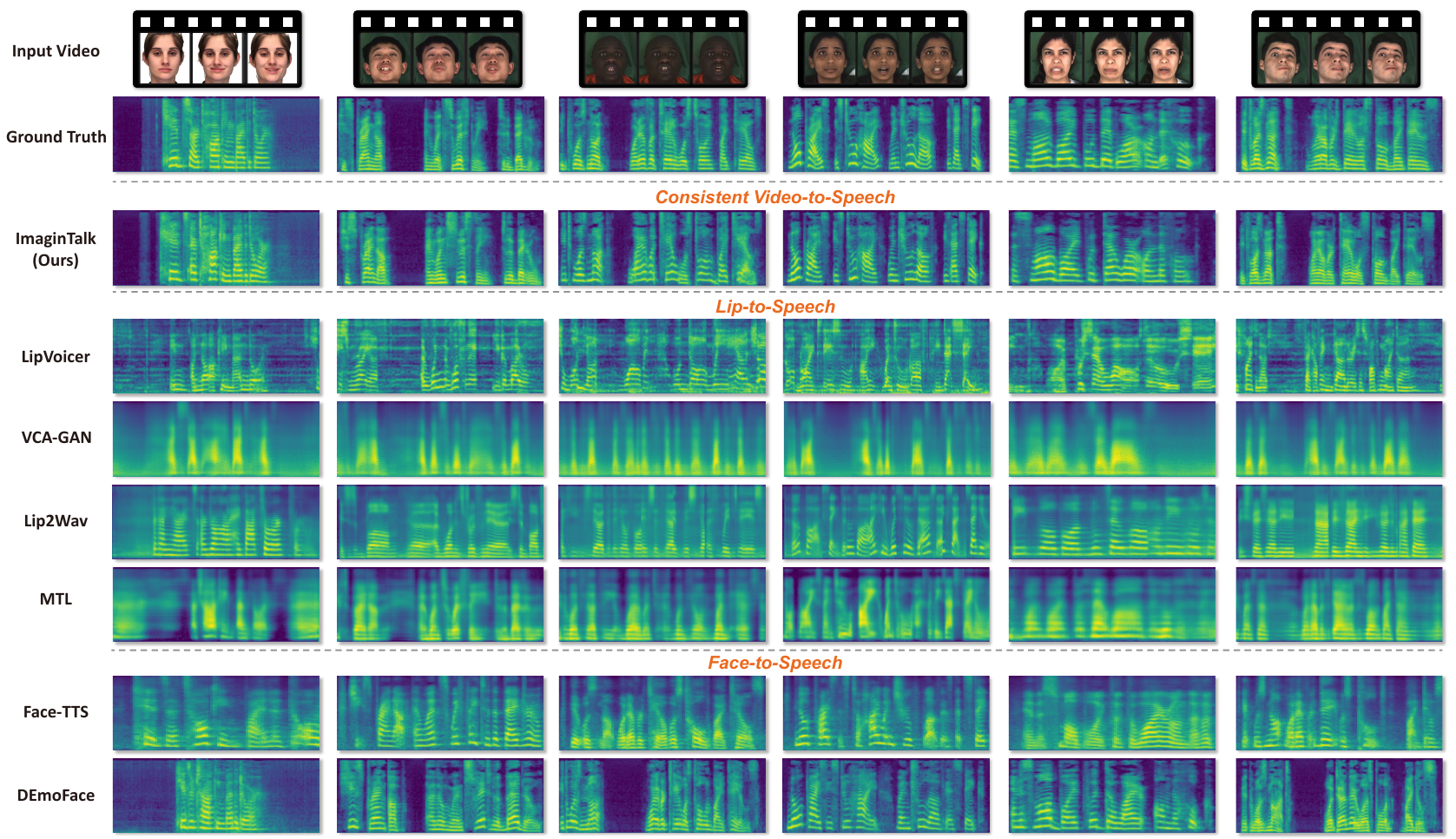}
    \caption{\textbf{Speech qualitative results on unseen speaker.} Zoom in for more details.}
    \label{fig:mel}
\end{figure*}

\paragraph{Implementation details.}\quad
For the speech tokenization, we employ a pre-trained RVQ-based codec from MaskGCT~\cite{maskgct:journals/corr/abs-2409-00750}, and adopt a log-linear noise schedule $\sigma(t)$~\cite{SEDD:conf/icml/LouME24} for the diffusion process.  where the expectation of the number of masked tokens is linear with time $t$. 
For our Style-DiT and discrete lip aligner,  we build DiT blocks upon DiT architecture~\cite{DiT:conf/iccv/PeeblesX23}, employ visual-temporal encoders inspired by SyncVSR~\cite{syncvsr:journals/corr/abs-2406-12233}, and incorporate RoBERTa based on vq-wav2vec~\cite{vqwav2vec:conf/iclr/BaevskiSA20}. 
During training, we use the AdamW optimizer~\cite{adamw/LoshchilovH19} with a learning rate of 1e-4, batch size 32. 
The total number of iterations is 300k. 
During inference, we use the Euler sampler to conduct the reverse process with 64 steps, and employ enhanced predictor-free guidance~\cite{demoface/abs-2502-01046} for multi-conditional guidance.

\subsection{Quantitative Evaluation}

We compare \methodname with state-of-the-art (SOTA) vision-guided methods, categorized into three paradigms based on task type: (\textbf{i}) Multimodal-to-speech (M2S)~\cite{visualvoicecloning/ChenTQZLW22,visualvoicecloning/Cong0QZWWJ0H23}, which introduces reference audio, text, and vision informtaion to tailor timbre, semantic, and prosody, respectively; (\textbf{ii}) Face-to-speech (F2S)~\cite{FaceTTS:conf/icassp/LeeCC23,demoface/abs-2502-01046}, which derives timbre and semantics directly from visual and text conditions; and  (\textbf{iii}) Lip-to-speech (L2S)~\cite{VCAGAN_lip:conf/nips/KimHR21,lipvoicer:conf/iclr/YeminiSBGF24}, which reconstructs semantics solely from visual conditions, while some methods~\cite{MTL_lip:conf/icassp/KimHR23,lip2wav:conf/cvpr/PrajwalMNJ20} additionally introduce audio cues to refine timbre.

\paragraph{Objective evaluation.}\quad
As shown in Tab.~\ref{tab:main}, compared with L2S methods without audio cues, we achieve 23\% and 56\% improvements in terms of EmoSim and SpkSim, reflecting the great ability to enhance emotion and timbre solely from the vision cues. We can also estimate the most precise F0 contour with a great ability in prosody dynamics modeling, and achieve a relative 33\% improvement in MCD that indicates minimal acoustic difference with the target speech. 
Furthermore, we introduce the audio-guided {ImaginTalk}$^*$ by replacing the face condition with the target audio features, which gains greater improvements than MTL and Lip2Wav in acoustic modeling by a large margin. 
While MTL achieves a better lip-synchronization through utilizing over 10 times the data, \methodname achieves comparable lip-sync accuracy while simultaneously improving speech naturalness and expressiveness. 
Notably, compared with M2S and F2S methods, our \methodname even outperforms them in terms of expressiveness and synchronization,
These results demonstrate that \methodname bridges the cross-modal gap using only heterogeneous vision features.

\begin{table}[t]
\setlength\tabcolsep{5.0pt}
\small
\centering
\begin{tabular}{rccc}
\toprule
Methods & $\text{MOS}_\text{nat}\uparrow$ & $\text{MOS}_\text{exp}\uparrow$ & $\text{MOS}_\text{syn}\uparrow$ \\ \hline
\specialrule{0em}{1.5pt}{1.5pt}
Ground Truth   & 3.72$\pm$0.38 & 3.81$\pm$0.37 & 3.62$\pm$0.54   \\
\hdashline
\specialrule{0em}{1.5pt}{1.5pt}
DEmoFace~\cite{demoface/abs-2502-01046}   & 3.31$\pm$0.51 & 3.51$\pm$0.14 & 2.10$\pm$0.84   \\
Lip2Wav~\cite{lip2wav:conf/cvpr/PrajwalMNJ20}   & 1.77$\pm$0.58 & 2.20$\pm$0.38 & 2.73$\pm$0.42   \\
LipVoicer~\cite{lipvoicer:conf/iclr/YeminiSBGF24}   & 2.07$\pm$0.63 & 2.71$\pm$0.47 & 3.07$\pm$0.20   \\
\specialrule{0em}{1.pt}{1.0pt}
\hline
\specialrule{0em}{1.pt}{1.0pt}
\methodname (\textbf{ours})   & \textbf{3.62$\pm$0.36} & \textbf{3.65$\pm$0.33} & \textbf{3.67$\pm$0.41} \\
\bottomrule
\end{tabular}
\caption{\textbf{Subjective evaluation} on speech naturalness, expressiveness, and synchronization, compared with other SOTA methods. \label{tab:userstudy}}
\vspace{-4.5mm}
\end{table}

\begin{figure*}[htbp]
    \centering
    \includegraphics[width=1.0\linewidth]{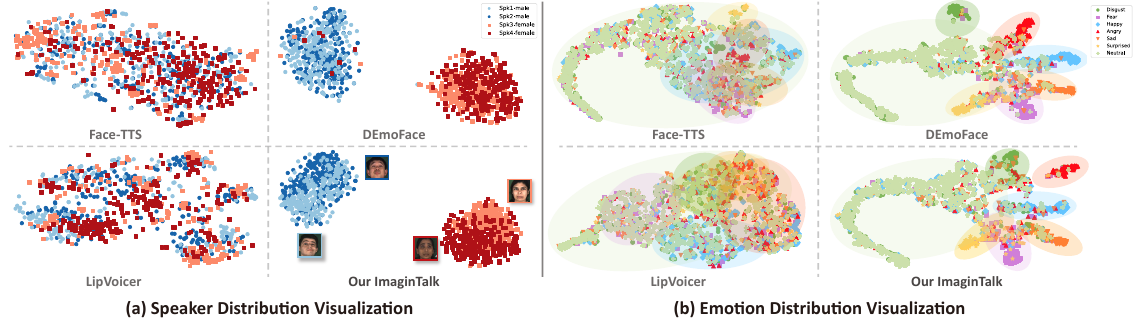}
    \vspace{-6mm}
    \caption{\textbf{t-SNE visualizations} of d-vectors and emotion2vec features. Each color represents a different speaker or emotion.}
    \label{fig:TSNE}
\end{figure*}

\paragraph{Subjective evaluation.}\quad
We further conduct the subjective evaluation with 15 participants, to compare our \methodname with SOTA methods.
Specifically, we introduce five mean opinion scores (MOS) with rating scores from 1 to 5 in 0.5 increments, including $\text{MOS}_\text{nat}$, $\text{MOS}_\text{exp}, \text{MOS}_\text{syn}$ for speech naturalness, expressiveness, and lip-synchronization. We randomly generate 10 samples from the test set. 
The scoring results of the user study are presented in Tab.~\ref{tab:userstudy}, demonstrating that \methodname achieves a distinct advantage over SOTA methods across all metrics, particularly surpassing 5\%in $\text{MOS}_\text{syn}$ with the ground truth. 
Furthermore, compared to F2S method, while achieving better acoustic modeling performance, we significantly enhance lip-synchronization with 31\% improvement. 
Compared to L2S methods without text-dependence, we generate high-quality and expressive speech using only visual cues, highlighting our ability to achieve cross-modal speech alignment.

\begin{table*}[thbp]
\setlength\tabcolsep{5.0pt}
\small
\centering
\begin{tabular}{lrrrrrrr}
\toprule
 Variants & EmoSim$\uparrow$ & SpkSim$\uparrow$ & RMSE$\downarrow$ &MCD$\downarrow$  & WER(\%)$\downarrow$ & LSE-C$\uparrow$ & LSE-D$\downarrow$\\ \hline
\specialrule{0em}{1.5pt}{1.5pt}
w/o RoBERTa   & 0.71 & 0.53  & 96.10 & 8.29  & 67.90 & 3.17 & 8.31 \\ 
w/o Error Detector   & 0.74 & \textbf{0.61}  & 88.94 & 7.75 & 67.50   & 3.38 & 8.12 \\ 
w/o Dual-adaLN  & \textbf{0.75} & 0.58  & 92.53 & 8.27  & 65.24 & 3.27 & 8.24 \\ 
\specialrule{0em}{1.pt}{1.0pt}
\hline
\specialrule{0em}{1.pt}{1.0pt}
\methodname (\textbf{ours}) & 0.74 & \textbf{0.61}  & \textbf{88.62} & \textbf{7.73} & \textbf{64.66} & \textbf{3.46} & \textbf{8.03} \\
\bottomrule
\end{tabular}
\caption{\textbf{Ablation studies.} ``Variants''refers to the ablation variants, indicating the utilization of an embedding layer instead of RoBERTa, removal of error detector, and replacing Dual-adaLN with vanilla adaLN, respectively. \label{tab:ablation}}
\vspace{-1mm}
\end{table*}

\subsection{Qualitative Results}
\paragraph{Qualitative spectrogram comparisons.}\quad 
As shown in Fig.~\ref{fig:mel}, we compare text-free L2S and text-dependent F2S methods. For L2S methods, we observe severe over-smoothing or acoustic artifacts, leading to significant degradation in speech quality and limiting their practical value. For F2S methods, while they can generate higher-quality speech with the aid of reference text, they struggle to align with the video, resulting in poor lip synchronization. In contrast, our method generates speech with richer acoustic details and precise lip synchronization, benefiting from our cross-modal diffusion process in the discrete space, which effectively addresses the one-to-many mapping issues.

\paragraph{Visualization of acoustic feature distribution.}\quad
To evaluate the effectiveness of acoustic modeling in terms of timbre (\ie,~speaker identity) and prosody (\ie,~emotion), we employ t-SNE~\cite{T-SNE} to visualize the distribution of d-vectors~\cite{dvector:journals/corr/LiWZZ15b} and emotion2vec~\cite{emo2vec:conf/acl/MaZYLGZ024} features extracted from the synthesized speech.
As shown in Fig.~\ref{fig:TSNE}(a), Face-TTS and LipVoicer exhibit poor speaker clustering, failing to distinguish identities and genders effectively, which results in the entanglement of different attributes. Although DEmoFace achieves partial clustering, some instances remain misaligned across genders. In contrast, our \methodname effectively distinguishes different speakers while achieving stronger gender-discriminative properties.
Furthermore, in Fig.~\ref{fig:TSNE}(b), methods that explicitly model prosody, such as DEmoFace and our \methodname, exhibit well-clustered emotional distributions, whereas other methods show significant overlap across different emotions. These results highlight the effectiveness of explicitly incorporating emotional cues in enhancing prosody modeling.

\subsection{Ablation Studies}
\paragraph{Ablation on lip aligner.}\quad 
To explore the impact of each component in the lip aligner, as shown in Tab.~\ref{tab:ablation}, we first replace RoBERTa with a learnable embedding layer without semantic prior to map tokens to semantic features, which affects other conditions to guide the denoising process, resulting in performance decline. Moreover, we observe that the removal of the error detector results in a degradation of WER while other metrics remain unaffected, indicating that the error detector effectively leverages masked language modeling of the RoBERTa to refine misaligned tokens.

\paragraph{Ablation on Dual-adaLN.}\quad 
To demonstrate the effectiveness of the Dual-adaLN, we utilize the vanilla adaLN of DiT~\cite{DiT:conf/iccv/PeeblesX23} and replace the temporal embedding with utterance-level emotional embedding combined with global style s an acoustic guidance. As shown in Tab.~\ref{tab:ablation}, although pooling dynamic emotions into an utterance-level style can enhance the emotional similarity, it struggles to model prosody dynamics with a 4\% relative decrease in terms of RMSE and results in entangling with other conditions with a further 3\% drop in SpkSim. The results highlight the effectiveness of Dual-adaLN.  

\section{Conclusion}
\label{sec:conclusion} 

We propose a novel cross-modal discrete diffusion framework to explore a challenging \taskname task, achieving unified cross-modal speech generation with semantics, timbre, and emotional prosody from visual cues simultaneously. 
Both quantitative and qualitative evaluations demonstrate that the proposed method can significantly improve expressiveness and reconstruct rich acoustic details, while achieving great lip-synchronization. 
For future work, while we have achieved good lip-reading using only 23 hours of data, we would further scale up the datasets with more diverse speakers and a larger vocabulary covering miscellaneous scenarios. 
Lastly, given the sensitivity of personal facial and auditory data, we emphasize the necessity of obtaining user consent for using published models, ensuring ethical application while safeguarding individual privacy and rights.

{
    \small
    \bibliographystyle{ieeenat_fullname}

\begin{thebibliography}{52}
        \providecommand{\natexlab}[1]{#1}
        \providecommand{\url}[1]{\texttt{#1}}
        \expandafter\ifx\csname urlstyle\endcsname\relax
          \providecommand{\doi}[1]{doi: #1}\else
          \providecommand{\doi}{doi: \begingroup \urlstyle{rm}\Url}\fi
        
        \bibitem[Afouras et~al.(2018)Afouras, Chung, and Zisserman]{LRS3/abs-1809-00496}
        Triantafyllos Afouras, Joon~Son Chung, and Andrew Zisserman.
        \newblock {LRS3-TED:} a large-scale dataset for visual speech recognition.
        \newblock \emph{CoRR}, abs/1809.00496, 2018.
        
        \bibitem[Ahn et~al.(2024)Ahn, Park, Park, Choi, and Kim]{syncvsr:journals/corr/abs-2406-12233}
        Youngjin Ahn, Jungwoo Park, Sangha Park, Jonghyun Choi, and Kee{-}Eung Kim.
        \newblock {SyncVSR}: {Data}-efficient visual speech recognition with end-to-end crossmodal audio token synchronization.
        \newblock In \emph{{Annu. Conf. Int. Speech Commun. Assoc.}}, 2024.
        
        \bibitem[Austin et~al.(2021)Austin, Johnson, Ho, Tarlow, and van~den Berg]{D3PM:conf/nips/AustinJHTB21}
        Jacob Austin, Daniel~D. Johnson, Jonathan Ho, Daniel Tarlow, and Rianne van~den Berg.
        \newblock Structured denoising diffusion models in discrete state-spaces.
        \newblock In \emph{{Adv. Neural Inform. Process. Syst.}}, pages 17981--17993, 2021.
        
        \bibitem[Baevski et~al.(2020)Baevski, Schneider, and Auli]{vqwav2vec:conf/iclr/BaevskiSA20}
        Alexei Baevski, Steffen Schneider, and Michael Auli.
        \newblock vq-wav2vec: {Self}-supervised learning of discrete speech representations.
        \newblock In \emph{{Int. Conf. Learn. Represent.}}, 2020.
        
        \bibitem[Chen et~al.(2022)Chen, Tan, Qi, Zhou, Li, and Wu]{visualvoicecloning/ChenTQZLW22}
        Qi Chen, Mingkui Tan, Yuankai Qi, Jiaqiu Zhou, Yuanqing Li, and Qi Wu.
        \newblock {V2C:} {Visual} voice cloning.
        \newblock In \emph{{IEEE Conf. Comput. Vis. Pattern Recog.}}, pages 21210--21219, 2022.
        
        \bibitem[Choi et~al.(2023)Choi, Hong, and Ro]{diffv2s:conf/iccv/ChoiHR23}
        Jeongsoo Choi, Joanna Hong, and Yong~Man Ro.
        \newblock {DiffV2S}: {Diffusion}-based video-to-speech synthesis with vision-guided speaker embedding.
        \newblock In \emph{{Int. Conf. Comput. Vis.}}, pages 7778--7787, 2023.
        
        \bibitem[Chung and Zisserman(2016)]{syncnet:conf/accv/ChungZ16a}
        Joon~Son Chung and Andrew Zisserman.
        \newblock Out of time: Automated lip sync in the wild.
        \newblock In \emph{{ACCV. Int. Worksh. }}, pages 251--263, 2016.
        
        \bibitem[Cong et~al.(2023)Cong, Li, Qi, Zha, Wu, Wang, Jiang, Yang, and Huang]{visualvoicecloning/Cong0QZWWJ0H23}
        Gaoxiang Cong, Liang Li, Yuankai Qi, Zheng{-}Jun Zha, Qi Wu, Wenyu Wang, Bin Jiang, Ming{-}Hsuan Yang, and Qingming Huang.
        \newblock Learning to dub movies via hierarchical prosody models.
        \newblock In \emph{{IEEE Conf. Comput. Vis. Pattern Recog.}}, pages 14687--14697, 2023.
        
        \bibitem[de~Mira et~al.(2022)de~Mira, Haliassos, Petridis, Schuller, and Pantic]{SVTS:conf/interspeech/MiraHPSP22}
        Rodrigo Schoburg~Carrillo de Mira, Alexandros Haliassos, Stavros Petridis, Bj{\"{o}}rn~W. Schuller, and Maja Pantic.
        \newblock {SVTS:} scalable video-to-speech synthesis.
        \newblock In \emph{{Annu. Conf. Int. Speech Commun. Assoc.}}, pages 1836--1840, 2022.
        
        \bibitem[D{\'{e}}fossez et~al.(2023)D{\'{e}}fossez, Copet, Synnaeve, and Adi]{encodec:journals/tmlr/DefossezCSA23}
        Alexandre D{\'{e}}fossez, Jade Copet, Gabriel Synnaeve, and Yossi Adi.
        \newblock High fidelity neural audio compression.
        \newblock \emph{Trans. Mach. Learn. Res.}, 2023, 2023.
        
        \bibitem[Deng et~al.(2022)Deng, Guo, Yang, Xue, Kotsia, and Zafeiriou]{arcface:journals/pami/DengGYXKZ22}
        Jiankang Deng, Jia Guo, Jing Yang, Niannan Xue, Irene Kotsia, and Stefanos Zafeiriou.
        \newblock Arcface: Additive angular margin loss for deep face recognition.
        \newblock \emph{{IEEE} Trans. Pattern Anal. Mach. Intell.}, 44\penalty0 (10):\penalty0 5962--5979, 2022.
        
        \bibitem[Deng et~al.(2023)Deng, Wu, Qiu, Luo, and Chen]{mixganTTS:journals/access/DengWQLC23}
        Yan Deng, Ning Wu, Chengjun Qiu, Yangyang Luo, and Yan Chen.
        \newblock Mixgan-tts: Efficient and stable speech synthesis based on diffusion model.
        \newblock \emph{{IEEE} Access}, 11:\penalty0 57674--57682, 2023.
        
        \bibitem[Goto et~al.(2020)Goto, Onishi, Saito, Tachibana, and Mori]{Face2Speech:conf/interspeech/GotoOSTM20}
        Shunsuke Goto, Kotaro Onishi, Yuki Saito, Kentaro Tachibana, and Koichiro Mori.
        \newblock {Face2Speech}: {Towards} multi-speaker text-to-speech synthesis using an embedding vector predicted from a face image.
        \newblock In \emph{{Annu. Conf. Int. Speech Commun. Assoc.}}, pages 1321--1325, 2020.
        
        \bibitem[Gulati et~al.(2020)Gulati, Qin, Chiu, Parmar, Zhang, Yu, Han, Wang, Zhang, Wu, et~al.]{conformer}
        Anmol Gulati, James Qin, Chung-Cheng Chiu, Niki Parmar, Yu Zhang, Jiahui Yu, Wei Han, Shibo Wang, Zhengdong Zhang, Yonghui Wu, et~al.
        \newblock Conformer: Convolution-augmented transformer for speech recognition.
        \newblock In \emph{{Annu. Conf. Int. Speech Commun. Assoc.}}, 2020.
        
        \bibitem[Guo et~al.(2023)Guo, Du, Chen, and Yu]{emodiff:conf/icassp/GuoDCY23}
        Yiwei Guo, Chenpeng Du, Xie Chen, and Kai Yu.
        \newblock {Emodiff}: {Intensity} controllable emotional text-to-speech with soft-label guidance.
        \newblock In \emph{{IEEE Conf. Acoust. Speech Signal Process.}}, pages 1--5, 2023.
        
        \bibitem[Hayashi et~al.(2017)Hayashi, Tamamori, Kobayashi, Takeda, and Toda]{RMSEf0:conf/asru/HayashiTKTT17}
        Tomoki Hayashi, Akira Tamamori, Kazuhiro Kobayashi, Kazuya Takeda, and Tomoki Toda.
        \newblock An investigation of multi-speaker training for wavenet vocoder.
        \newblock In \emph{{IEEE Autom. Speech Recognit. Understanding Worksh.}}, pages 712--718, 2017.
        
        \bibitem[He et~al.(2016)He, Zhang, Ren, and Sun]{resnet}
        Kaiming He, Xiangyu Zhang, Shaoqing Ren, and Jian Sun.
        \newblock Deep residual learning for image recognition.
        \newblock In \emph{{IEEE Conf. Comput. Vis. Pattern Recog.}}, pages 770--778, 2016.
        
        \bibitem[Huang et~al.(2022)Huang, Lam, Wang, Su, Yu, Ren, and Zhao]{fastdiff:conf/ijcai/HuangL0S00Z22}
        Rongjie Huang, Max W.~Y. Lam, Jun Wang, Dan Su, Dong Yu, Yi Ren, and Zhou Zhao.
        \newblock Fastdiff: {A} fast conditional diffusion model for high-quality speech synthesis.
        \newblock In \emph{{Proc. Int. Joint Conf. Artif. Intell.}}, pages 4157--4163, 2022.
        
        \bibitem[Jang et~al.(2024)Jang, Kim, Ahn, Kwak, Yang, Ju, Kim, Kim, and Chung]{facespeak:conf/cvpr/JangKAKYJKKC24}
        Youngjoon Jang, Ji{-}Hoon Kim, Junseok Ahn, Doyeop Kwak, Hongsun Yang, Yooncheol Ju, Ilhwan Kim, Byeong{-}Yeol Kim, and Joon~Son Chung.
        \newblock Faces that speak: {Jointly} synthesising talking face and speech from text.
        \newblock In \emph{{IEEE Conf. Comput. Vis. Pattern Recog.}}, pages 8818--8828, 2024.
        
        \bibitem[Kang et~al.(2023)Kang, Han, and Yang]{facestylespeech:journals/corr/abs-2311-05844}
        Minki Kang, Wooseok Han, and Eunho Yang.
        \newblock Face-stylespeech: Improved face-to-voice latent mapping for natural zero-shot speech synthesis from a face image.
        \newblock \emph{CoRR}, abs/2311.05844, 2023.
        
        \bibitem[Kelly(2011)]{kelly2011reversibility}
        Frank~P Kelly.
        \newblock \emph{Reversibility and stochastic networks}.
        \newblock Cambridge University Press, 2011.
        
        \bibitem[Kim et~al.(2024)Kim, Kim, and Chung]{lip2speech_AAAI:conf/aaai/KimKC24}
        Ji{-}Hoon Kim, Jaehun Kim, and Joon~Son Chung.
        \newblock Let there be sound: {Reconstructing} high quality speech from silent videos.
        \newblock In \emph{{AAAI Conf. Artif. Intell.}}, pages 2759--2767, 2024.
        
        \bibitem[Kim et~al.(2021)Kim, Hong, and Ro]{VCAGAN_lip:conf/nips/KimHR21}
        Minsu Kim, Joanna Hong, and Yong~Man Ro.
        \newblock Lip to speech synthesis with visual context attentional {GAN}.
        \newblock In \emph{{Adv. Neural Inform. Process. Syst.}}, pages 2758--2770, 2021.
        
        \bibitem[Kim et~al.(2023)Kim, Hong, and Ro]{MTL_lip:conf/icassp/KimHR23}
        Minsu Kim, Joanna Hong, and Yong~Man Ro.
        \newblock Lip-to-speech synthesis in the wild with multi-task learning.
        \newblock In \emph{{IEEE Conf. Acoust. Speech Signal Process.}}, pages 1--5, 2023.
        
        \bibitem[Lee et~al.(2023)Lee, Chung, and Chung]{FaceTTS:conf/icassp/LeeCC23}
        Jiyoung Lee, Joon~Son Chung, and Soo{-}Whan Chung.
        \newblock Imaginary voice: Face-styled diffusion model for text-to-speech.
        \newblock In \emph{{IEEE Conf. Acoust. Speech Signal Process.}}, pages 1--5, 2023.
        
        \bibitem[Li et~al.(2015)Li, Wang, Zhang, and Zheng]{dvector:journals/corr/LiWZZ15b}
        Lantian Li, Dong Wang, Zhiyong Zhang, and Thomas~Fang Zheng.
        \newblock Deep speaker vectors for semi text-independent speaker verification.
        \newblock \emph{CoRR}, abs/1505.06427, 2015.
        
        \bibitem[Li et~al.(2024)Li, Cheng, He, Peng, and Hauptmann]{mmtts_emo:journals/corr/abs-2404-18398}
        Xiang Li, Zhi{-}Qi Cheng, Jun{-}Yan He, Xiaojiang Peng, and Alexander~G. Hauptmann.
        \newblock {MM-TTS:} {A} unified framework for multimodal, prompt-induced emotional text-to-speech synthesis.
        \newblock \emph{CoRR}, abs/2404.18398, 2024.
        
        \bibitem[Liu et~al.(2019)Liu, Ott, Goyal, Du, Joshi, Chen, Levy, Lewis, Zettlemoyer, and Stoyanov]{roberta:journals/corr/abs-1907-11692}
        Yinhan Liu, Myle Ott, Naman Goyal, Jingfei Du, Mandar Joshi, Danqi Chen, Omer Levy, Mike Lewis, Luke Zettlemoyer, and Veselin Stoyanov.
        \newblock {RoBERTa}: {A} robustly optimized {BERT} pretraining approach.
        \newblock \emph{CoRR}, abs/1907.11692, 2019.
        
        \bibitem[Livingstone and Russo(2018)]{RAVDESS}
        Steven~R Livingstone and Frank~A Russo.
        \newblock The ryerson audio-visual database of emotional speech and song ({RAVDESS}): {A} dynamic, multimodal set of facial and vocal expressions in north american english.
        \newblock \emph{PLOS ONE}, 13\penalty0 (5):\penalty0 e0196391, 2018.
        
        \bibitem[Loshchilov and Hutter(2019)]{adamw/LoshchilovH19}
        Ilya Loshchilov and Frank Hutter.
        \newblock Decoupled weight decay regularization.
        \newblock In \emph{{Int. Conf. Learn. Represent.}}, 2019.
        
        \bibitem[Lou et~al.(2024)Lou, Meng, and Ermon]{SEDD:conf/icml/LouME24}
        Aaron Lou, Chenlin Meng, and Stefano Ermon.
        \newblock Discrete diffusion modeling by estimating the ratios of the data distribution.
        \newblock In \emph{{Int. Conf. on Mach. Learn.}}, 2024.
        
        \bibitem[Ma et~al.(2024)Ma, Zheng, Ye, Li, Gao, Zhang, and Chen]{emo2vec:conf/acl/MaZYLGZ024}
        Ziyang Ma, Zhisheng Zheng, Jiaxin Ye, Jinchao Li, Zhifu Gao, Shiliang Zhang, and Xie Chen.
        \newblock emotion2vec: Self-supervised pre-training for speech emotion representation.
        \newblock In \emph{{Findings Proc. Annu. Meeting Assoc. Comput. Linguistics}}, pages 15747--15760. Association for Computational Linguistics, 2024.
        
        \bibitem[Mao et~al.(2023)Mao, Xu, Yin, Chang, Nie, and Huang]{posterv2:journals/corr/abs-2301-12149}
        Jiawei Mao, Rui Xu, Xuesong Yin, Yuanqi Chang, Binling Nie, and Aibin Huang.
        \newblock {POSTER} {V2:} {A} simpler and stronger facial expression recognition network.
        \newblock \emph{CoRR}, abs/2301.12149, 2023.
        
        \bibitem[Meng et~al.(2022)Meng, Choi, Song, and Ermon]{ConcreteScoreMatch:conf/nips/MengCSE22}
        Chenlin Meng, Kristy Choi, Jiaming Song, and Stefano Ermon.
        \newblock Concrete score matching: Generalized score matching for discrete data.
        \newblock In \emph{{Adv. Neural Inform. Process. Syst.}}, 2022.
        
        \bibitem[Ou et~al.(2024)Ou, Nie, Xue, Zhu, Sun, Li, and Li]{RADD:journals/corr/abs-2406-03736}
        Jingyang Ou, Shen Nie, Kaiwen Xue, Fengqi Zhu, Jiacheng Sun, Zhenguo Li, and Chongxuan Li.
        \newblock Your absorbing discrete diffusion secretly models the conditional distributions of clean data.
        \newblock \emph{CoRR}, abs/2406.03736, 2024.
        
        \bibitem[Peebles and Xie(2023)]{DiT:conf/iccv/PeeblesX23}
        William Peebles and Saining Xie.
        \newblock Scalable diffusion models with transformers.
        \newblock In \emph{{Int. Conf. Comput. Vis.}}, pages 4172--4182, 2023.
        
        \bibitem[Prajwal et~al.(2020)Prajwal, Mukhopadhyay, Namboodiri, and Jawahar]{lip2wav:conf/cvpr/PrajwalMNJ20}
        K.~R. Prajwal, Rudrabha Mukhopadhyay, Vinay~P. Namboodiri, and C.~V. Jawahar.
        \newblock Learning individual speaking styles for accurate lip to speech synthesis.
        \newblock In \emph{{IEEE Conf. Comput. Vis. Pattern Recog.}}, pages 13793--13802, 2020.
        
        \bibitem[Radford et~al.(2023)Radford, Kim, Xu, Brockman, McLeavey, and Sutskever]{whisper/RadfordKXBMS23}
        Alec Radford, Jong~Wook Kim, Tao Xu, Greg Brockman, Christine McLeavey, and Ilya Sutskever.
        \newblock Robust speech recognition via large-scale weak supervision.
        \newblock In \emph{{Int. Conf. on Mach. Learn.}}, pages 28492--28518, 2023.
        
        \bibitem[Schroff et~al.(2015)Schroff, Kalenichenko, and Philbin]{facenet:conf/cvpr/SchroffKP15}
        Florian Schroff, Dmitry Kalenichenko, and James Philbin.
        \newblock Facenet: {A} unified embedding for face recognition and clustering.
        \newblock In \emph{{IEEE Conf. Comput. Vis. Pattern Recog.}}, pages 815--823, 2015.
        
        \bibitem[Subakan et~al.(2021)Subakan, Ravanelli, Cornell, Bronzi, and Zhong]{sepformer/SubakanRCBZ21}
        Cem Subakan, Mirco Ravanelli, Samuele Cornell, Mirko Bronzi, and Jianyuan Zhong.
        \newblock Attention is all you need in speech separation.
        \newblock In \emph{{IEEE Conf. Acoust. Speech Signal Process.}}, pages 21--25, 2021.
        
        \bibitem[Sun et~al.(2023)Sun, Yu, Dai, Schuurmans, and Dai]{SCDDM:conf/iclr/SunYDSD23}
        Haoran Sun, Lijun Yu, Bo Dai, Dale Schuurmans, and Hanjun Dai.
        \newblock Score-based continuous-time discrete diffusion models.
        \newblock In \emph{{Int. Conf. Learn. Represent.}}, 2023.
        
        \bibitem[Van~der Maaten and Hinton(2008)]{T-SNE}
        Laurens Van~der Maaten and Geoffrey Hinton.
        \newblock Visualizing data using t-{SNE}.
        \newblock \emph{J. Mach. Learn. Res.}, 9\penalty0 (11), 2008.
        
        \bibitem[Wan et~al.(2018)Wan, Wang, Papir, and L{\'{o}}pez{-}Moreno]{GE2E/WanWPL18}
        Li Wan, Quan Wang, Alan Papir, and Ignacio L{\'{o}}pez{-}Moreno.
        \newblock Generalized end-to-end loss for speaker verification.
        \newblock In \emph{{IEEE Conf. Acoust. Speech Signal Process.}}, pages 4879--4883, 2018.
        
        \bibitem[Wang et~al.(2020)Wang, Wu, Song, Yang, Wu, Qian, He, Qiao, and Loy]{MEAD:conf/eccv/WangWSYWQHQL20}
        Kaisiyuan Wang, Qianyi Wu, Linsen Song, Zhuoqian Yang, Wayne Wu, Chen Qian, Ran He, Yu Qiao, and Chen~Change Loy.
        \newblock {MEAD:} {A} large-scale audio-visual dataset for emotional talking-face generation.
        \newblock In \emph{{Eur. Conf. Comput. Vis.}}, pages 700--717, 2020.
        
        \bibitem[Wang et~al.(2018)Wang, Stanton, Zhang, Skerry{-}Ryan, Battenberg, Shor, Xiao, Jia, Ren, and Saurous]{WER/WangSZRBSXJRS18}
        Yuxuan Wang, Daisy Stanton, Yu Zhang, R.~J. Skerry{-}Ryan, Eric Battenberg, Joel Shor, Ying Xiao, Ye Jia, Fei Ren, and Rif~A. Saurous.
        \newblock Style tokens: {Unsupervised} style modeling, control and transfer in end-to-end speech synthesis.
        \newblock In \emph{{Int. Conf. on Mach. Learn.}}, pages 5167--5176, 2018.
        
        \bibitem[Wang et~al.(2024)Wang, Zhan, Liu, Zeng, Guo, Zheng, Zhang, Zhang, and Wu]{maskgct:journals/corr/abs-2409-00750}
        Yuancheng Wang, Haoyue Zhan, Liwei Liu, Ruihong Zeng, Haotian Guo, Jiachen Zheng, Qiang Zhang, Shunsi Zhang, and Zhizheng Wu.
        \newblock {MaskGCT}: Zero-shot text-to-speech with masked generative codec transformer.
        \newblock \emph{CoRR}, abs/2409.00750, 2024.
        
        \bibitem[Wu et~al.(2024)Wu, Li, Liu, and Yang]{DCTTS:conf/icassp/WuLLY24}
        Zhichao Wu, Qiulin Li, Sixing Liu, and Qun Yang.
        \newblock {DCTTS:} {Discrete} diffusion model with contrastive learning for text-to-speech generation.
        \newblock In \emph{{IEEE Conf. Acoust. Speech Signal Process.}}, pages 11336--11340. {IEEE}, 2024.
        
        \bibitem[Xue et~al.(2023)Xue, Liu, He, Tan, Liu, Lin, and Zhao]{foundationTTS:journals/corr/abs-2303-02939}
        Ruiqing Xue, Yanqing Liu, Lei He, Xu Tan, Linquan Liu, Edward Lin, and Sheng Zhao.
        \newblock Foundationtts: Text-to-speech for {ASR} customization with generative language model.
        \newblock \emph{CoRR}, abs/2303.02939, 2023.
        
        \bibitem[Yang et~al.(2023)Yang, Yu, Wang, Wang, Weng, Zou, and Yu]{diffsound:journals/taslp/YangYWWWZY23}
        Dongchao Yang, Jianwei Yu, Helin Wang, Wen Wang, Chao Weng, Yuexian Zou, and Dong Yu.
        \newblock {Diffsound}: {Discrete} diffusion model for text-to-sound generation.
        \newblock \emph{{IEEE} {ACM} Trans. Audio Speech Lang. Process.}, 31:\penalty0 1720--1733, 2023.
        
        \bibitem[Ye et~al.(2025)Ye, Cao, and Shan]{demoface/abs-2502-01046}
        Jiaxin Ye, Boyuan Cao, and Hongming Shan.
        \newblock Emotional face-to-speech.
        \newblock \emph{CoRR}, abs/2502.01046, 2025.
        
        \bibitem[Yemini et~al.(2024)Yemini, Shamsian, Bracha, Gannot, and Fetaya]{lipvoicer:conf/iclr/YeminiSBGF24}
        Yochai Yemini, Aviv Shamsian, Lior Bracha, Sharon Gannot, and Ethan Fetaya.
        \newblock {LipVoicer}: {Generating} speech from silent videos guided by lip reading.
        \newblock In \emph{{Int. Conf. Learn. Represent.}}, 2024.
        
        \bibitem[Zheng et~al.(2024)Zheng, Tu, Xiao, and Xu]{Srcodec:conf/icassp/ZhengTXX24}
        Youqiang Zheng, Weiping Tu, Li Xiao, and Xinmeng Xu.
        \newblock Srcodec: {Split}-residual vector quantization for neural speech codec.
        \newblock In \emph{{IEEE Conf. Acoust. Speech Signal Process.}}, pages 451--455, 2024.
        
        \end{thebibliography}

}

\end{document}